\documentclass[10pt,english,oneside,twocolumn,a4paper]{article} 
\usepackage{graphicx}


\usepackage{graphicx}

\usepackage{xcolor,colortbl}

\renewcommand{\thefootnote}{\alph{footnote}}
\newcommand\freefootnote[1]{%
  \let\thefootnote\relax%
  \footnotetext{#1}%
  \let\thefootnote\svthefootnote%
}

\usepackage[utf8]{inputenc}
\usepackage{times}
\fontfamily{ptm}\selectfont
\usepackage{tabularx}
\usepackage{ragged2e}
\usepackage[singlelinecheck=false]{caption}
\usepackage[T1]{fontenc}
\usepackage[numbers]{natbib}
\usepackage{amsmath}
\usepackage{amssymb}
\usepackage[switch]{lineno}
\usepackage{multirow}

\usepackage{makecell}
\usepackage{booktabs}

\usepackage{subfig}
\usepackage{graphicx} 
\usepackage{subcaption}
\usepackage{babel}
\RequirePackage[blocks]{authblk}
\makeatletter
\let\ps@plain\ps@empty
\def\@xivpt{14pt}
\def\@sect#1#2#3#4#5#6[#7]#8{%
  \ifnum #2<2
    \null\par\vskip-15pt
  \fi
  \ifnum #2>\c@secnumdepth 
    \let\@svsec\@empty
  \else
    \refstepcounter{#1}%
    \protected@edef\@svsec{%
      \ifnum #2<4
        \hb@xt@10mm{\csname the#1\endcsname}\relax
      \else
        \hb@xt@12mm{\csname the#1\endcsname}\relax
      \fi}%
  \fi
  \@tempskipa #5\relax
  \ifdim \@tempskipa>\z@
    \begingroup
      #6{%
        \@hangfrom{\hskip #3\relax\@svsec}%
          \interlinepenalty \@M #8\@@par}%
    \endgroup
    \csname #1mark\endcsname{#7}%
    \addcontentsline{toc}{#1}{%
      \ifnum #2>\c@secnumdepth \else  
        \protect\numberline{\csname the#1\endcsname}%
      \fi 
      #7}%
  \else
    \def\@svsechd{%
      #6{\hskip #3\relax
      \@svsec #8}%
      \csname #1mark\endcsname{#7}%
      \addcontentsline{toc}{#1}{%
        \ifnum #2>\c@secnumdepth \else
          \protect\numberline{\csname the#1\endcsname}%
        \fi
        #7}}%
  \fi
  \@xsect{#5}}
\renewcommand\LARGE{\@setfontsize\LARGE{16}{20}}
\def\abstract#1{\def\@abstract{#1}}
\def\abstractEn#1{\def\@abstractEn{#1}}
\def\titleEn#1{\def\@titleEn{#1}}
\def\@maketitle{%
  \newpage
  \null
  \let \footnote \thanks
    {\LARGE\bfseries\RaggedRight \@title \par}%
    \vskip 1\baselineskip%
    {\normalsize
      \@author\par}%
    \vskip 2\baselineskip%
    \vskip \baselineskip%
    {\section*{Abstract}
      \@abstract}%
  \par
  \vskip 3\baselineskip}

\renewcommand\section{\@startsection {section}{1}{\z@}%
                                   {-3.5ex \@plus -1ex \@minus -.2ex}%
                                   {\baselineskip}%
                                   {\normalfont\Large\bfseries\RaggedRight}}
\renewcommand\subsection{\@startsection{subsection}{2}{\z@}%
                                     {\baselineskip}%
                                     {1ex}%
                                     {\normalfont\large\bfseries\RaggedRight}}
\renewcommand\subsubsection{\@startsection{subsubsection}{3}{\z@}%
                                     {1\baselineskip}%
                                     {3bp}%
                                     {\normalfont\normalsize\bfseries\RaggedRight}}
\renewcommand\paragraph{\@startsection{paragraph}{4}{\z@}%
                                    {1\baselineskip\@plus1ex \@minus.2ex}%
                                    {3bp}%
                                    {\normalfont\normalsize\RaggedRight}}
\renewcommand\subparagraph{\@startsection{subparagraph}{5}{\parindent}%
                                       {3.25ex \@plus1ex \@minus .2ex}%
                                       {-1em}%
                                      {\normalfont\normalsize\bfseries\RaggedRight}}
\parindent\p@
\parskip\z@
\DeclareCaptionLabelSeparator{enskip}{\enskip}
\makeatother

\title{Can SAR improve RSVQA performance?}

\author[a,b]{Lucrezia Tosato}
\author[a]{Sylvain Lobry}
\author[b]{Flora Weissgerber} 
\author[a]{Laurent Wendling}
\affil[a]{LIPADE, Université Paris Cité, 75006 Paris, France}
\affil[b]{ONERA, Traitement de l’information et systèmes, 91123 Palaiseau, France}

\abstract{
Remote sensing visual question answering (RSVQA) 
has been involved in several research in recent years, leading to an increase in new methods. RSVQA automatically extracts information from satellite images, so far only optical, and a question to automatically search for the answer in the image and provide it in a textual form. 
In our research, we study whether Synthetic Aperture Radar (SAR) images can be beneficial to this field. We divide our study into three phases which include classification methods and VQA. In the first one, we explore the classification results of SAR alone and investigate the best method to extract information from SAR data. Then, we study the combination of SAR and optical data. In the last phase, we investigate how SAR images and a combination of different modalities behave in RSVQA compared to a method only using optical images. We conclude that adding the SAR modality leads to improved performances, although further research on using SAR data to automatically answer questions is needed as well as more balanced datasets.}
\begin{document}

\maketitle

\section{Introduction}
\freefootnote{The experiments conducted in this study were performed using HPC/AI resources provided by GENCI-IDRIS (Grant 2023-AD011012735R2).}
Large amounts of remote sensing imagery are now readily available thanks to the efforts of the public and private sectors. A strong example is Sentinel satellites launched in 2014 as part of the European Union's Copernicus program. This mission offers free access to different types of images (including multi-spectral and Synthetic Aperture Radar (SAR)) with large spatial coverage and a short revisit time. 

The information contained within the satellite images could be used by citizens or journalists to verify events, conflicts or the climate crisis, in almost real-time. 
In addition, the ability to accurately interpret SAR images not only provides the opportunity to acquire information during cloudy days or at night, ensuring a continuous flow of information, but it also offers unique insights into material composition and texture.

However, it is difficult to extract information from remote sensing images. 
This interpretation is usually done by experts and often involves manual work, which becomes a limiting factor as the amount of data produced increases. 
Interpretation is even more difficult in SAR due to its higher degree of complexity. 
SAR data interpretation challenges are due to geometric distortions (foreshortening, layover, shadow) and speckle noise. They also result from complex radar backscatter influenced by surface roughness, dielectric properties, and radar geometry. 
\begin{figure}[ht!]
    \centering
    \includegraphics[width=1\columnwidth]{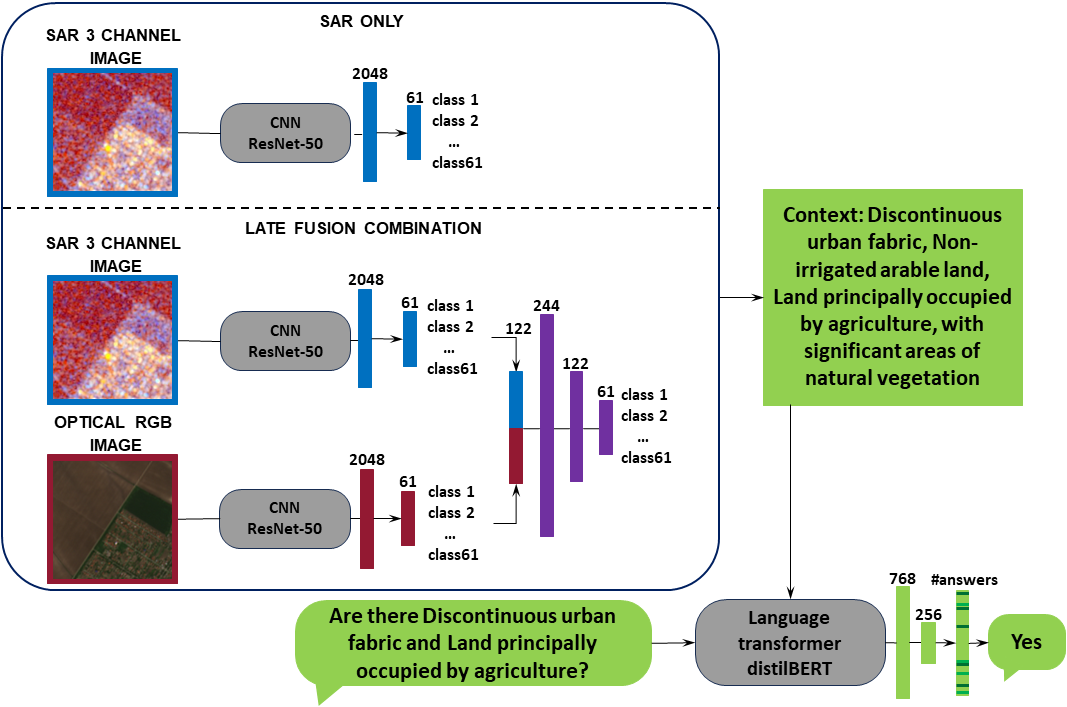}
    \caption{Outline of the proposed method. We propose to extract land cover classes from SAR data or from SAR and optical data. These classes are then used as an input, with the question, to a language model that predicts the answer. }\label{pipeline}
\end{figure}
In this work, we propose to combine optical and SAR information in order to be able to automatically answer questions in natural language about the satellite images.


This task is called Remote Sensing Visual Question Answering (RSVQA) and originates from Visual Question Answering (VQA), whose objective is to predict a precise natural language response to an open-ended question related to an image~\cite{antol2015vqa}.
The RSVQA task was first introduced by~\cite{lobry2020rsvqa}. In this work, the authors provide a first dataset and a method that extracts separately textual and visual features from the question and image respectively. The features are then merged, resulting in a final vector used to make the prediction, i.e. answer the question asked. 

This method, despite promising results with optical images, does not address the problem of explainability. Indeed, all the process is done in one step, preventing the observation of intermediate results.
In PROMPT-RSVQA~\cite{Chappuis_2022_CVPR} this problem is addressed by dividing the process into two different phases. In the first, the model aims at identifying classes that are relevant to answer the question from the image. In the second step, features are extracted from the question and merged into a transformer with the names of the classes identified in the image. The output of the transformer is a textual response to the question. By dividing the method into two steps, it is possible to study the classes detected in the images in more detail and better understand the origin of potential prediction errors.

In the field of RSVQA, only optical images have been used so far. Due to the complexity of SAR, the interaction between SAR images and text only starts to be studied.  
Some studies on SAR captioning have addressed this interaction.  
SAR is not good in describing the size of targets and in counting objects~\cite{zhao2022exploring}. The modality has good results in outlining the relative position of the target object, such as whether it is close to other objects or classes and their density description(e.g. "few", "a lot", "many", \ldots).

To combine optical and SAR images different fusions were studied, using a modified version of hybrid pansharpening~\cite{dalla2015challenges} with SAR and optical images, halfway fusion~\cite{zhu2022sar}, early fusion~\cite{sumbul2019bigearthnet} and late fusion.
Early fusion merges data from different sources before processing, halfway fusion combines results from separate sources, while late fusion processes data separately and then combines the results afterwards.
A comparison between early fusion and late fusion in different modalities was made in ~\cite{wagner2016multispectral} concluding that late fusion leads to better results. 

This work divides the study into 3 research questions leading to a method for RSVQA from SAR and optical images. We propose a method that follows the framework of PROMPT-RSVQA~\cite{Chappuis_2022_CVPR}, illustrated in Figure~\ref{pipeline}: we first extract land cover classes from the image modality (i.e. a multi-label classification task). The predicted class names are then put as input to a language model alongside the question to predict an answer.

The first research question investigates how to maximize classification results from SAR images. The second phase explores how to merge SAR images and optical ones to improve the SAR-only and optical-only classification results. 
The last step studies how using only SAR or a combination of SAR and optical as an input to a VQA model improves the results compared to only using optical data as an input.

To answer these research questions, the datasets that we use in this study are presented in section 2. We describe the proposed methods to answer the 3 research questions in section 3, and we describe the evaluation methods in section 4. We present and discuss the results in section 5.

\section{Data}
\begin{figure}[t!]
    \centering
    \includegraphics[width=0.8\columnwidth]{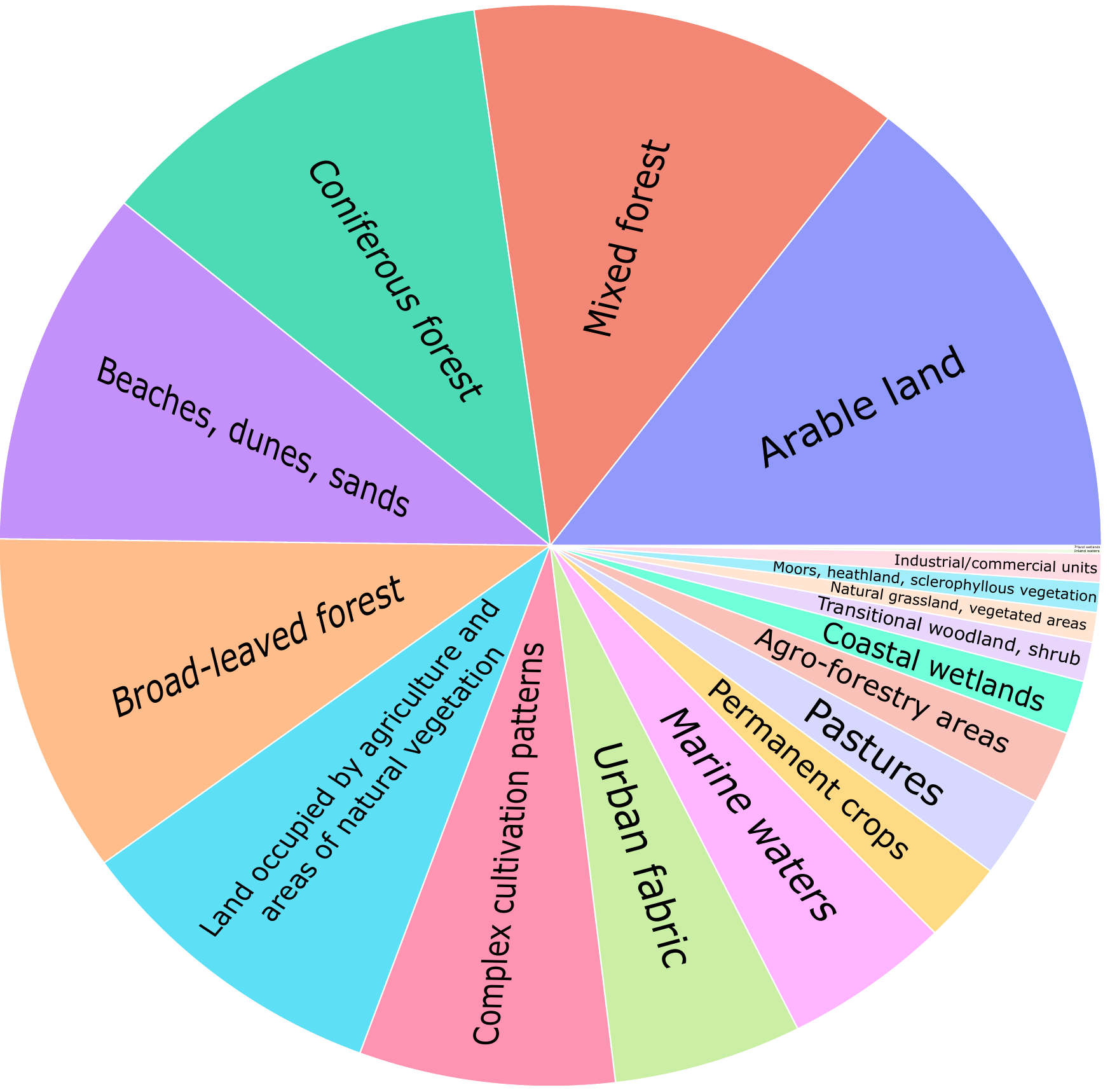}
    \caption{Distribution of the 19 classes from the BEN-MM dataset.}\label{19classes}
\end{figure}

We use two datasets: BEN-MM for the land cover classification and RSVQAxBEN for the VQA downstream task.\\
\textbf{BigEarthNet-Multi Modality dataset~\cite{sumbul2021bigearthnet}} (BEN-MM) contains 590'326 patches for both Sentinel 1 (VV and VH in dB) and Sentinel 2 (with 12 channels). It adds to the BigEarthNet~\cite{sumbul2019bigearthnet} (BEN) dataset, which only contained Sentinel 2 images.
The dataset is acquired over 10 different European countries providing the acquisition information, and associates images with the CORINE Land Cover (CLC) map of 2018.  
In this nomenclature, the land cover classes are presented on three increasingly specific levels L1, L2 and L3 for a total of 64 classes. 
In our work, we consider two nomenclatures extracted from the CLC taxonomy. The first one (named \textbf{BEN-MM split}, the same as in~\cite{sumbul2021bigearthnet}) groups different classes and eliminates classes at the L3 level, creating a total of 19 categories shown in Figure~\ref{19classes}.
In this setting, we follow the random dataset split into training, validation and testing sets made by~\cite{sumbul2019bigearthnet}.    

\begin{figure}[t!]
    \centering
    \includegraphics[width=1\columnwidth]{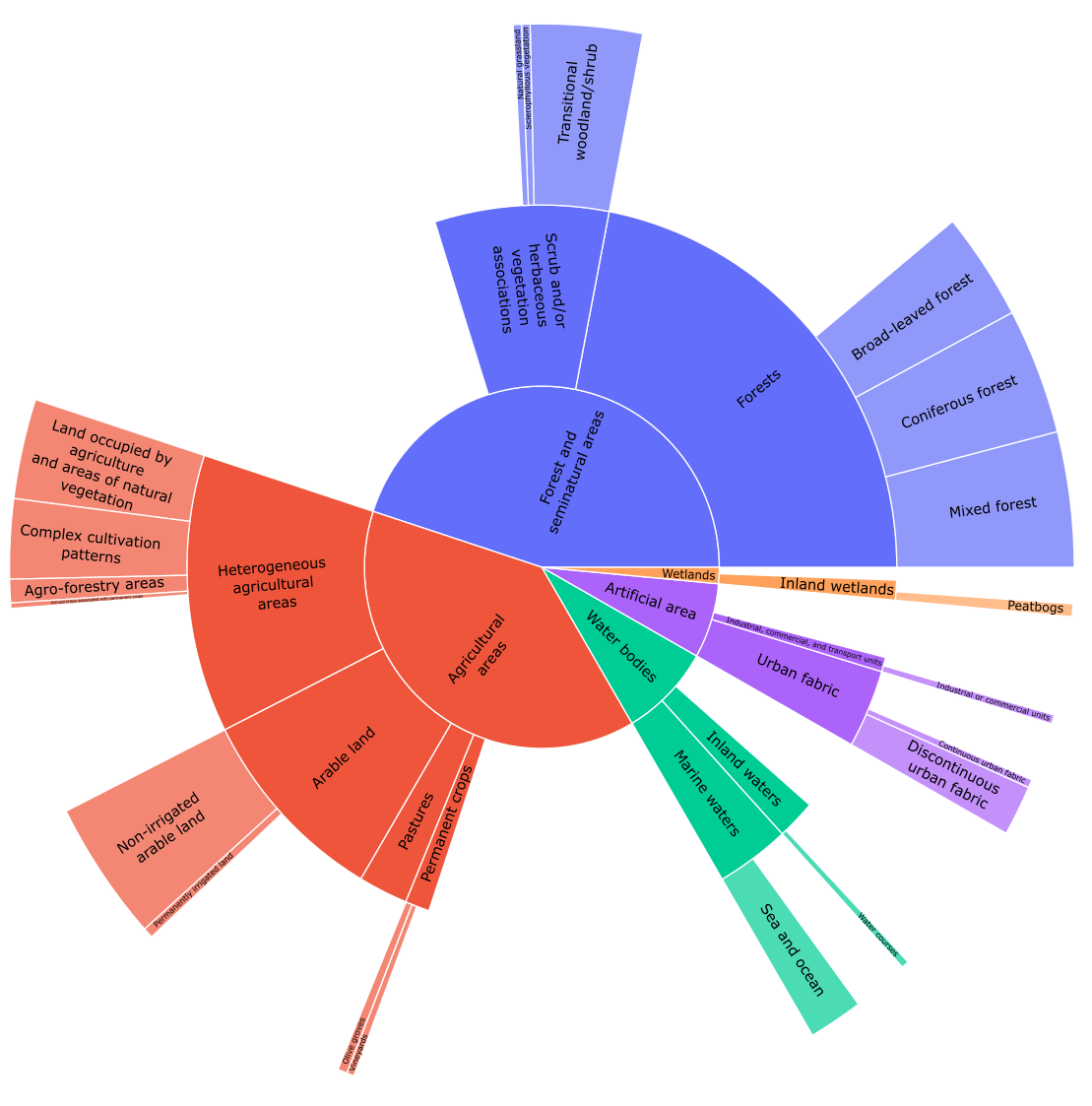}
    \caption{Distribution of the 36 classes present classes out of the original 61. The inner, middle, and outside circles represent L1, L2 and L3 classes respectively.}\label{61classes}
\end{figure}

In the second nomenclature (named \textbf{RSVQA split}, the same as in~\cite{lobry2021rsvqa}), we use the 64 CLC classes. We count as a single label the two classes with the same name at different hierarchy levels (water bodies and pastures) and eliminate the categories of 'Glaciers and perpetual snow', leading to a total of 61 classes. Figure~\ref{61classes} presents the 36 most present classes out of the 61, underlining the increasing imbalance in the various classes. 
\\
\textbf{RSVQAxBEN dataset~\cite{lobry2021rsvqa}} contains Sentinel 2 patches of BEN and the related questions and answers. The questions are based on each image level of the CLC labels, i.e., the terrain features. The question type is in 80.7\% of the cases a \textit{yes/no question}, the remaining are \textit{Land cover questions} requiring as an answer the list of classes present.
In \textit{yes/no questions}, conjunctions such as 'and', 'or' are present in 72.3\% of the questions. In 27.1\% of the questions, two conjunctions are found.

For each patch, 25 questions and answers are present, creating a total of 14'758'150 image/question/answer triplets. 
The dataset split is divided according to the spatial location of the image, this choice allows the models to avoid biases due to geographical location. 
\section{Method}
\subsection{The polarization ratio's impact on multi-label classification}
In our first step, we want to understand if it's better to use 2 (VV and VH) or 3 channels (VV, VH, and the ratio) to predict land cover classes. Using two polarizations instead of one has been proven to be more useful in separating vegetation from artificial targets~\cite {lee2001quantitative}. We want to investigate whether the network alone is capable of creating a linear combination of the two other channels, or whether it needs to be included. 

In the first case, the 2 normalised and saturated (from a cumulative histogram study of all images) VV and VH channels are used.  
The first 2 channels are the same as in the 2 channel case, and the third one is their ratio which has been shown to help classify the volume from the surface~\cite{park2008integration}. Using images in dB, our ratio is a subtraction of the two channels. 
\subsection{Comparison between early fusion and late fusion }
In the second step, we compare a late fusion strategy with the early fusion applied in~\cite{sumbul2021bigearthnet}. The early fusion of~\cite{sumbul2021bigearthnet} concatenates the optical and SAR channels together as an input to the neural network. The late fusion, on the other hand combines the results of the classification, after the feature extraction step. 
For both these settings, we predict the 19 land cover classes. We use a ResNet-50~\cite{he2016deep} model pre-trained on the ImageNet~\cite{deng2009imagenet} dataset. The split used is the one of~\cite{sumbul2021bigearthnet} and the loss function is trained using class weights in the loss, defined as the inverse of the frequency of the apparition of a class. To adapt the architecture of ResNet-50 to multi-label classification, we replace the softmax layer with a sigmoid layer. 
These predictions are thresholded 
to optimize scores on the validation set.

Considering $x$ the number of classes, the late fusion concatenates the two vectors after the sigmoid function and inputs the $x\cdot2$ elements to a multi-layer perceptron. The output of this network is a vector containing $x$ elements, which represents the probabilities of the classes predicted using the combination of information from the two modalities.

\subsection{Impact of classification results for VQA}
In the third step, we study how SAR-only and the combination behave in VQA compared to optical-only. 
We propose 3 steps of our proposed method (A, B and C) to gradually switch from the setting used to compare the classification results of BEN-MM to the VQA results of PROMPT-RSVQA~\cite{Chappuis_2022_CVPR}.
\begin{table}[t!]
\begin{center}
{\small\begin{tabular}{|c|c|c|c|}\hline
& A & B & C \\ \hline
Classes & 61  & 61  & 61  \\ \hline
Dataset Split & BEN-MM  & RSVQA   & RSVQA  \\ \hline
Weighted loss & Yes  & Yes  & No  \\ \hline
\end{tabular}}
\end{center}
\caption{Description of the setting of the three steps (A, B, C) of the third research question. }\label{thirdstepABC}
\end{table}

The change of parameters in the steps is presented in Table~\ref{thirdstepABC}.
At the end of step C, we input the names of the detected classes (after thresholding the results) into a language model, following the setting of PROMPT-RSVQA~\cite{Chappuis_2022_CVPR}. The complete architecture using only SAR images and the combination is presented in Figure~\ref{pipeline}.

The language model uses the question and the textual description of the image as inputs to generate a feature vector. This vector effectively combines information from both the visual and textual modalities. Within our framework, we employ DistilBERT as the language model. DistilBERT is built upon an attention-based Transformer architecture~\cite{sanh2019distilbert}. It is a more compact iteration of the Bidirectional Encoder Representations from Transformers (BERT)~\cite{devlin2018bert}. 
The weights of the language model are the same as in PROMPT-RSVQA~\cite{Chappuis_2022_CVPR}.

\section{Performance evaluation}\label{ssec:vqa_metric}
The first two research questions are related to classification and are therefore evaluated using classification metrics. The last question links classification and VQA, so other metrics related to context and VQA are proposed.  
\subsection{Classification}
\textbf{Macro and Micro Score:} Macro score is calculated by averaging the per-class scores: each class is treated regardless of its frequency in the dataset.  Micro score is calculated by aggregating the contributions of each individual instance in the dataset: each instance is treated regardless of its class label. Given $N$ the total number of classes, $s_{1,2..N}$ the scores for classes 1 to N, $w_{1,2..N}$ the weight given to a class and  $o_{1,2..N}$ the number of actual occurrences of the class in the dataset, the formula for the calculation of the scores is shown in equation~\ref{eq1}.
\begin{equation} 
\begin{aligned}
M = & s_1*w_1+s_2*w_2+...s_N*w_N \\
w_{1,2...N} = &
\begin{cases}
\frac{1}{N} & \text{if } M = \text{Macro} \\
o_{1,2..N} & \text{if } M = \text{Micro}
\end{cases} 
\end{aligned}
\label{eq1} 
\end{equation}

\textbf{$\boldsymbol{F_{\beta}}$-score:} 
Precision (\textit{P}) represents the proportion of positive predictions that are correct, while recall (\textit{R}) denotes the proportion of actual positives that are identified correctly. The $F_{\beta}$-score is calculated as a weighted harmonic mean of  \textit{P} and \ \textit{R}, with the parameter $\beta$\ controlling the balance between these two important aspects of model performance.
It is commonly used to evaluate the performance of a binary classification model. With $\beta=1$ indicating equal weight, $\beta<1$ favouring precision, and
$\beta>1$ favouring recall 
\begin{equation}{F_\beta = (1+ \beta^2) \cdot \frac{P \cdot R}{\beta^2 \cdot P+R}} \label{eq2} \end{equation} 

\subsection{From classification to VQA}
\textbf{Match Ratio (MR):} 
The MR computes the fraction of correctly classified samples, i.e. the samples whose predicted labels exactly correspond to the ground truth labels. This gives the following for \textit{Q} samples:
\begin{equation}
    MR = \frac{1}{Q}\sum_{i=1}^Q I(y_i=\hat{y}_i)\,,\label{eq3}
\end{equation}
where $y_i$ represents the actual labels for the $i_{th}$ sample, while $\hat{y}_i$ represents the labels predicted by the model for the same sample, expressed as a one-hot encoded vector. The function $I$ evaluates whether the actual labels match exactly with the predicted labels for each sample. This function returns 1 for an exact match and 0 otherwise. \\
\textbf{Hamming Distance (HD): }
The HD is defined as the number of classes that have a different prediction than the ground truth. It is defined for \textit{Q} samples and \textit{N} land cover categories as:
\begin{equation}
    HD = \frac{1}{Q}\sum_{i=1}^Q \sum_{j=1}^N I(y_{ij}=\hat{y}_{ij})\label{eq4}\,,
\end{equation}
where $y_{ij}$ and $\hat{y}_{ij}$ represent the prediction and ground truth of a specific value of a land cover class (0 or 1) and $I$ is the indicator function. 
\subsection{VQA}
To evaluate VQA results, we define the percentage of correct answers as the accuracy. The global (overall samples) and per type of question (“Yes/No” or “Land cover” subsets) accuracies are provided. With this metric, all mistakes have the same weight regardless of their similarity to the ground truth.

\section{Results and discussion} 

The results we obtained aim to answer 3 research questions. The first investigates which SAR channel(s) to use for better classification results. The second step explores how to merge SAR images and optical ones to improve SAR-only and optical-only results. The last step study how SAR-only and the combination behave in VQA compared to optical-only.

\subsection{The polarization ratio's impact on multi-label classification}
 Table~\ref{23chaSAR} shows that using three channels (VV, VH and ratio) instead of 2 (VV and VH) can be useful for classification. In particular, out of the 19 classes, 14 have better results with 3 channels, leading to an overall increase of F2 Macro score close to 2\%. The classes that benefit most are \textit{Beaches, dunes} and \textit{Moors, heathland} with an increase of 8.82\% and 7.79\% respectively. This could be explained by the volumetric information given by the ratio of the two polarizations, but more in-depth studies should be carried out to confirm this hypothesis. 
\begin{table}[t!]
\begin{center}
{\small\begin{tabular}{|l|c|}\hline
&F2 Macro score\\ \hline
S1 2CH& 57.11\% \\ \hline
S1 3CH& \textbf{58.84\%} \\ \hline

\end{tabular}}
\end{center}
\caption{Classification results from Sentinel 1 with 2 (S1 2CH) and 3 (S1 3CH) channels. The best results are shown in bold.}\label{23chaSAR}
\end{table}
\subsection{Comparison between early fusion and late fusion}
In Table~\ref{fusion}, we present the classification results with one modality only, the early fusion (EF) used in~\cite{sumbul2021bigearthnet} and our proposed late fusion (LF). The late fusion of SAR and optical images improves results with respect to optical only in all the classes and has higher results than BEN-MM early fusion in 17 out of the 19 classes. 
On this dataset, early fusion and S2 have similar results: the absolute value of the average difference between the combined and optical results in~\cite{sumbul2021bigearthnet} is 5.70\%. In our findings, we obtain a difference of 11.32\%. This could be due to the fact that a single neural network for SAR and optical images does not offer enough capacity.
\begin{table}[t!]
\begin{center}
{\small\begin{tabular}{|l|c|}\hline
&F2 Macro score \\  \hline
S1 3CH&  58.84\% \\ \hline
S2 & 67.12\% \\ \hline
EF (BEN-MM~\cite{sumbul2021bigearthnet}) & 67.23\% \\ \hline
LF (Ours) &  \textbf{73.87\%} \\ \hline
\end{tabular}}
\end{center}
\caption{Classifications results with single modality models, early fusion and late fusion. The best results are shown in bold.}\label{fusion}
\end{table}
\subsection{Impacts of classification results for VQA} 
Our results on the last research question are presented in Table~\ref{promptsar} and ~\ref{classtot}, through the outcomes of optical (S2), 3 channels SAR (S1) and our late fusion (LF) classification. We present the results for the 3 steps defined in section~\ref{ssec:vqa_metric}. We move from the F2 macro score classification evaluation to the F1 micro and F1 score to compare our results to the one of PROMPT-RSVQA~\cite{Chappuis_2022_CVPR}.

\begin{table*}[t!]
\centering
\small
\begin{tabular}{|c|c|c|c|c|c|c|c|}
\hline
\multicolumn{2}{|c|}{} & F1Micro & HD & MR & GA & Y/N A & LC A\\
\specialrule{.2em}{.1em}{.1em}
\multirow{3}{*}{A} & S2 & 69.01\%  & 4.72 & 13.6\% & - & - & - \\
\cline{2-8}
& S1 & 61.44\%  & 6.21 & 6.9\% & - & - & -  \\
\cline{2-8}
& LF (Ours) & 81.88\%  & 3.07 & 20.0\% & - & - & -  \\
\specialrule{.2em}{.1em}{.1em}
\multirow{3}{*}{B} & S2 & 66.66\% & 4.75 & 12.0\% & 70.58\%
 & 81.72\% & 19.55\% \\
\cline{2-8}
& S1 & 65.07\% & 5.15 & 10.3\% & 70.14\% & 81.31\% & 18.98\% \\
\cline{2-8}
& LF (Ours) & 69.62\% & 4.01 & 12.8\% & 72.22\% & 83.48\% & 20.64\% \\
\specialrule{.2em}{.1em}{.1em}
\multirow{3}{*}{C}& PROMPT-RSVQA~\cite{Chappuis_2022_CVPR} (S2) & 74.60\% & 3.4 & 15.6\% & 75.40\% & \textbf{86.07\%} & 26.56\% \\
\cline{2-8}
& S1 & 67.26\% & 4.24 & 11.9\% & 71.78\% & 82.94\% & 20.64\% \\
\cline{2-8}
& LF (Ours) & 75.00\% & 3.35 & 14.3\% & \textbf{75.49\%} & \textbf{86.07\%} & \textbf{27.03\%} \\
\hline
\end{tabular}
\caption{Classification and VQA results calculated using the evaluation scores presented in Section \ref{ssec:vqa_metric}. The best results for the VQA task are shown in bold.  
 }\label{promptsar}
\end{table*}

\begin{table*}[t!]
\centering
\small
\begin{tabular}{|c|c||c|c|c||c||c|c|c|c|}
\hline
\multicolumn{2}{|c||}{} & \multicolumn{1}{c|}{Inland waters} & \multicolumn{1}{c|}{Broad-leaved forest} & \multicolumn{1}{c||}{Agriculture and Vegetation} & \multicolumn{1}{c||}{Forest and Natural areas} & \multicolumn{1}{c|}{Peatbogs} \\
\hline
\multicolumn{2}{|c||}{Frequency} & \multicolumn{1}{c|}{0.0152} & \multicolumn{1}{c|}{0.0317} & \multicolumn{1}{c||}{0.0293} & \multicolumn{1}{c||}{0.1492} & \multicolumn{1}{c|}{0.0037} \\
\specialrule{.2em}{.1em}{.1em}
\multirow{3}{*}{A} & S2 & 74.64\% & 69.10\% & 61.84\% & 73.94\% & 47.75\% \\
\cline{2-7}
& S1 & 77.04\% & 65.40\% & 58.94\% & 68.34\% & 32.87\% \\
\cline{2-7}
& LF (Ours) & \textbf{84.56\%} & \textbf{74.43\%} & \textbf{67.53\%}  & \textbf{93.76\%}& \textbf{71.70\%}\\
\specialrule{.2em}{.1em}{.1em}
\multirow{3}{*}{B} & S2 & 40.68\% & 17.13\% & 45.38\% & 98.51\% & 28.15\% \\
\cline{2-7}
& S1 & \textbf{79.55}\% & \textbf{17.91\%} & 38.81\% & \textbf{98.52\%} & 0.4\% \\
\cline{2-7}
& LF (Ours) & 79.02\% & 16.09\% & \textbf{47.67\%} & 98.38\% & \textbf{31.69\%} \\
\specialrule{.2em}{.1em}{.1em}
\multirow{3}{*}{C} & S2 & 74.92\% &14.64\% &45.45\% & 98.80\%& \textbf{40.32\%} \\
\cline{2-7}
& S1 & \textbf{80.71\%} & \textbf{17.74\%} &  46.93\% &  98.65\%&  0\%\\
\cline{2-7}
& LF (Ours) & 77.75\% & 15.10\% & \textbf{51.50\%}  &  \textbf{98.83\%} &  35.42\%\\
\hline
\end{tabular}
\caption{The F1 score results of some classes through all three stages and their dataset frequency. The best results per class and per phase are shown in bold.}
\label{classtot}
\end{table*}
We can see that the global accuracy (GA) calculated using the combination is better than that of optical images. Indeed while 
“Yes/No” accuracy (Y/N A) is the same, “Land cover” accuracy (LC A) improves.

The MR and HD scores indicate that by using the combination, the final predicted label is less frequently completely correct resulting in a lower match ratio.
On the other hand, fewer errors are present in the land cover prediction, as shown by a lower hamming distance than the optical images or SAR only. 
This implies fewer errors, but they are more widely distributed in the predictions.

An improvement of these metrics from step B to step C is present. Indeed, removing the weights from the loss improves the VQA results but makes classification strongly dependent on the frequency of the classes in the dataset. This is necessary to perform well in VQA, as the questions and answers concerning a class are proportional to the frequency of that class but worsen the improvement the combination brings.

The classification results show a deterioration of the results between steps A and B as B requires the model to be more generalisable. Only three classes have better combination results in step B than in A. These classes are \textit{Forest and seminatural areas}, \textit{Forests} and \textit{Mixed forest}. These classes are hierarchically one inside the other, and very easily generalisable as they are not specific to one place. 
Note that also in \textit{Inland waters} and \textit{Forest and Natural areas} classes, there is no drop in classification results for SAR, as shown in Table~\ref{classtot}. 
For \textit{Forest and Natural areas} the reason is the one just cited while for \textit{Inland waters} this could be due to the fact that the SAR backscattering of an inland water body is very low, and thus does not change with location.

The small improvement of the VQA metrics is due to the variable impact of the combination in the classification results of the different classes in step C. The behaviour of the classes can be divided into three groups as shown in Table~\ref{classtot}. The first group are classes in which the SAR classification is better than the optical classification and the combination has better results.
In this group there are 6 classes, among them\textit{Inland waters}, \textit{Broad-leaved forest} and \textit{Agriculture and Vegetation}.
The second group are classes in which the SAR classification is worse than the optical-only but the results of the combination are still better or equal. Here we find 35 classes, including \textit{Forest and Natural areas }. 
The final group are classes in which the SAR classification is lower and the combination gets worse. This group represents 1/3 of the total classes, of whom \textit{Peatbogs}. 

\begin{figure}[t!]
    \centering
    \includegraphics[width=0.99\columnwidth]{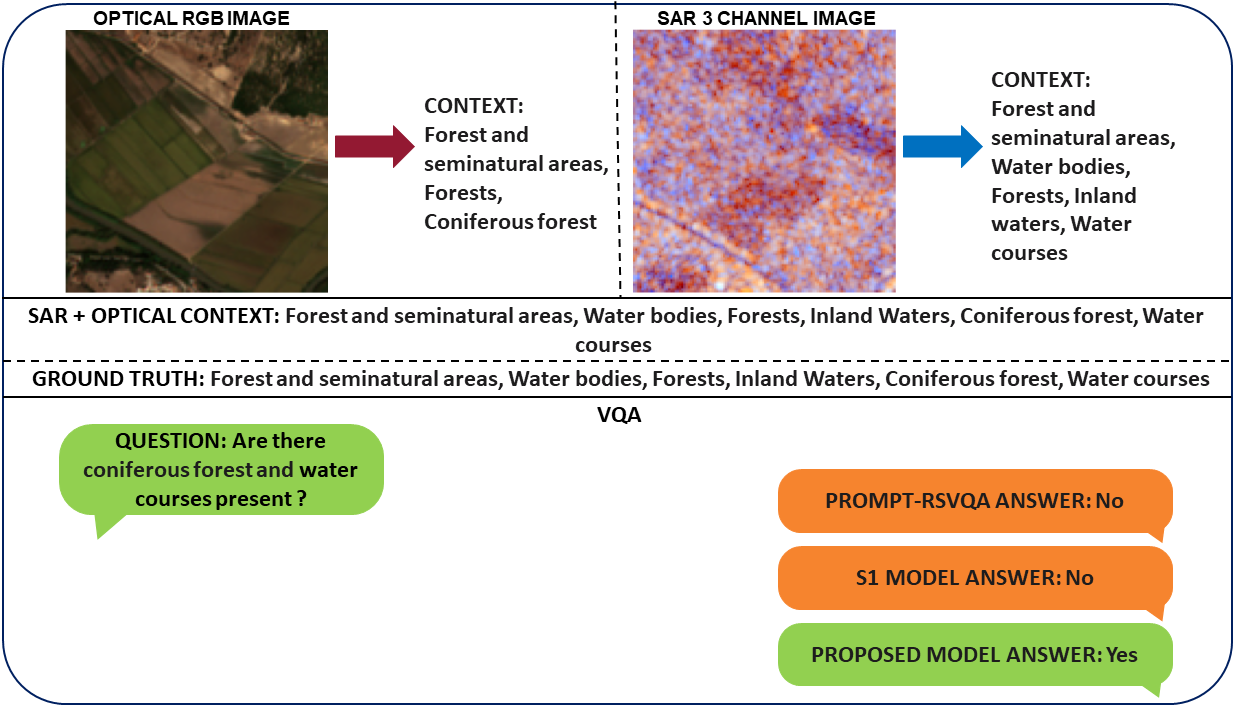}
    \caption{Comparison of SAR alone, optical alone and their late fusion combination in VQA.}\label{qa}
\end{figure}

Most of the water classes benefit from the use of SAR. However, these classes are under-represented in RSVQAxBEN giving a low overall impact. 
Also, almost all the wetlands classes show a deterioration in combination results.  More in-depth studies are needed on this topic.
Figure~\ref{qa} presents a qualitative result. In this example, the model using optical image alone does not predict the right classes to give the right answer. Indeed, the two classes of water are only identified by SAR. 

Combining the SAR information with the optical information makes the answer complete. 

\section{Conclusion}
Our article highlights the promising potential of SAR in the field of RSVQA. The study is divided into 3 phases. In the first, we show that the use of 3 channels in SAR classification is useful. In the second, we prove that a late fusion has better results than an early fusion in our dataset. 
In the third, we study in more detail the classification results of many classes in a highly unbalanced dataset and the translation of the classification results into VQA ones. 
From the proposed metrics, we deduce that the combination produces fewer errors, but more spread out in the land cover predictions. 
We show how generalisation leads to a decrease in performance, except for a few strongly generalisable classes. We show that by removing the weights from the loss the results in classification worsen while those in VQA improve.
Finally, we discuss the behaviour of the 61 classes, comparing the combination with the optical results alone, showing that there are 3 different groups.

To the best of our knowledge, our proposed method is able to leverage SAR data to obtain the best results on the RSVQAxBEN dataset.

We can conclude that in order to achieve similar combination behaviour in all three phases, it is necessary to apply data augmentation and domain adaptation to improve the generalisation of the classes. On the other hand, to maintain good results in classification and VQA, it is necessary to have balanced classes and answers in the dataset. 
Furthermore, the complexity of SAR in the context of VQA and its broader applicability in the field of explicability call for a comprehensive exploration. A crucial first step in this direction is SAR captions, which can clarify the direct relationship between textual information and SAR constructs. By delving into these aspects, valuable information can be gained that paves the way for improving the role of SAR in the understanding and interpretation of visual content.

\end{document}